\documentclass{bmvc2k}

\usepackage{graphicx}
\usepackage{amsmath,amssymb} 
\usepackage{bm}
\usepackage{booktabs}
\usepackage{rotating,multirow}
\usepackage{wrapfig}

\usepackage{tikz}
\usepackage{subfigure}

\title{Novel View Synthesis from Single Images via Point Cloud Transformation}

\addauthor{Hoang-An Le}{hoang-an.le@uva.nl}{1}
\addauthor{Thomas Mensink}{mensink@google.com}{2}
\addauthor{Partha Das}{p.das@uva.nl}{1}
\addauthor{Theo Gevers}{th.gevers@uva.nl}{1}

\addinstitution{
 Computer Vision Lab, \\
 University of Amsterdam\\
}
\addinstitution{
 Google Research, \\
 Amsterdam
}

\runninghead{Le \etal}{Novel View Synthesis via Point Cloud Transformation}

\def\eg{\emph{e.g}\bmvaOneDot} 
\def\ie{\emph{i.e}\bmvaOneDot} 
\def\cf{\emph{c.f}\bmvaOneDot}

\def\etal{\emph{et al}\bmvaOneDot}
\makeatother

\newcommand{\myparagraph}[1]{\textbf{#1}\quad}

\newcommand{\CR}[1]{{\color{black}#1}}

\begin{document}

\maketitle

\begin{abstract}
    In this paper the argument is made that for true novel view synthesis of objects, where the object can be synthesized from any viewpoint, an explicit 3D shape representation is desired. 
Our method estimates point clouds to capture the geometry of the object, which can be freely rotated into the desired view and then projected into a new image. 
This image, however, is sparse by nature and hence this coarse view is used as the input of an image completion network to obtain the dense target view.
The point cloud is obtained using the predicted pixel-wise depth map, estimated from a single RGB input image, combined with the camera intrinsics.
By using forward warping and backward warping between the input view and the target view, the network can be trained end-to-end without supervision on depth. 
The benefit of using point clouds as an explicit 3D shape for novel view synthesis is experimentally validated on the 3D ShapeNet benchmark. Source code and data are available at \url{https://github.com/lhoangan/pc4novis}

\end{abstract}

\graphicspath{{images/}}

\section{Introduction}

Novel view synthesis aims to infer the appearance of an object from unobserved points of view. 
The synthesis of unseen views of objects could be  important for image-based 3D object manipulation~\cite{Kholgade2014}, robot traversability~\cite{Hirose2019}, or 3D object reconstruction~\cite{Tatarchenko2016}. Generating a coherent view of unseen parts of an object requires a non-trivial understanding of the object's inherent properties such as (3D) geometry, texture, shading, and illumination.

Different algorithms make use of provided source images in different ways. Model-based approaches use similar-look open stock 3D models~\cite{Kholgade2014}, or through user interactive construction~\cite{Zheng2012,Chen2013,Rematas2017}.
Image-based
methods~\cite{Tatarchenko2016,Zhou2016,Park2017,Sun2018,Olszewski2019}
assume an underlying parametric model of object appearances conditioned on viewpoints and try to learn it using statistical frameworks. Despite their differences, both approaches use 3D
information in predicting object new views. The former imposes stronger
assumptions on the full 3D structure and shifts the paradigm to obtain the full 
models, 
while the latter captures the 3D information in latent space to cope with (self) occlusion. 

\begin{figure*}[t]
    \centering
    \includegraphics[width=.95\textwidth]{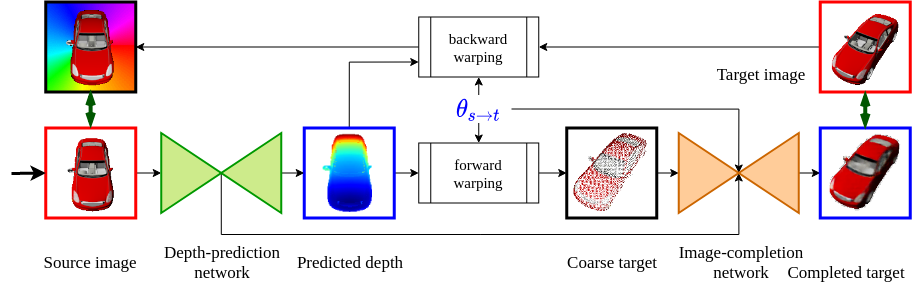}
    \caption{
    Overview of the proposed model for training and inference. 
    From a single input image, the pixel-wise depth map is predicted. The depth map is subsequently used to compute a coarse novel view (forward warping), and trained by making use of backward warping (from the target view back to the source view). The model is trained end-to-end.
    }
    \label{fig:pipeline}
    \vspace{-3mm}
\end{figure*}

The principle is that the generation of a new view of an object is composed of (1) relocating pixels in source images that will be visible to the corresponding positions in the target view, (2) removing the pixels that will be occluded, and (3) adding disoccluded pixels that are not seen in the source and will be visible in the target view~\cite{Park2017}. With the advance of convolution neural networks (CNNs) and generative adversarial networks (GANs), 
\cite{Zhou2016,Park2017,Sun2018} show that (1) and (2) can be done by
learning an appearance flow field that "flows" pixels from a source image to the
corresponding positions in the target view, and (3) can be done by a completion network with an adversarial loss.

In this paper, we leverage the explicit use of geometry information in synthesizing
novel views. We argue that (1) and (2) can be done in a straightforward manner by obtaining access to the geometry of the objects. The appearance flow~\cite{Zhou2016,Park2017,Sun2018} which associates pixels of the source view to their positions in the target view, is the projection of the 3D displacement of objects' points before and after transformation. Occluded object parts can be identified based on the orientation of the object surface normals and the view directions. The argument can also be extended for multiple input images. In this paper, we show that the geometry of an object provides an explicit and natural basis to the problem of novel view synthesis.

In contrast to geometry-based methods, the proposed approach does not require 3D supervision. The method predicts a depth map in a self-supervised manner by formulating the depth estimation problem in the context of novel view synthesis. The predicted depth is used to partly construct the target views and to assist the completion network.

The main contributions of this paper are:  
(1) a novel methodology for novel view synthesis using explicit transformations of estimated point clouds;   (2) an integrated model combining self-supervised monocular depth estimation and novel view synthesis, which can be trained end-to-end; 
(3) natural extensions to multi-view inputs and full point cloud reconstruction from a single image; and
(4) experimental benchmarking to validate the proposed method, which outperforms the current state-of-the art methods for novel view synthesis.

\section{Related Work}
\subsection{Geometry-based view synthesis}
\noindent\myparagraph{View synthesis via 3D models}
Full models (textured meshes or colored point clouds) of objects or scenes are constructed from multiple images taken from various
viewpoints
~\cite{Debevec1996,Seitz2006,Meshry2019} or are given and aligned interactively by users~\cite{Kholgade2014,Rematas2017}. The use of 3D models allows for extreme pose estimation, re-texturing and flexible (re-)lighting by applying rendering techniques~\cite{Nguyen2018,Meshry2019}. However, obtaining complete 3D models of objects or scenes is a challenging task in itself. Therefore, these approaches require additional user input to identify objects boundaries~\cite{Zheng2012,Chen2013}, select and align 3D models with image views~\cite{Kholgade2014,Rematas2017}, or use simple textured-mapped 3-planar billboard models~\cite{Hoiem2005}. In contrast, the proposed method makes use of objects partial point clouds constructed from a given source view and does not require a predefined (explicit) 3D model.

\noindent\myparagraph{View synthesis via depth}
Methods using 3D models assume a coherent structure between the desired
objects and the obtained 3D models~\cite{Chen2013,Kholgade2014}. Synthesis using depth obtains an intermediate representation from depth information. The intermediate representation captures hidden surfaces from one or multiple viewpoints. \cite{Zitnick2004} proposes to use layered depth images, 
\cite{Flynn2016} creates 3D plane sweep volumes by projecting 
images onto target viewpoints at different depths, \cite{Zhou2018stereo} uses
multi-plane images at fix-distances to the camera, and \cite{Choi2019} estimates
depth probability volumes to leverage depth uncertainty in occluded regions.

In contrast, the proposed method estimates depth directly from monocular views to partially construct the target views. Self-supervised depth estimation using deep neural networks using photometric re-projection consistency has been researched by several
authors~\cite{Garg2016,Zhou2017,Godard2017,Godard2019,Johnston2019}. In this paper, we train a self-supervised depth prediction network with novel view
synthesis in an end-to-end system.

\subsection{Image-based view synthesis}

Requiring explicit geometrical structures of objects or scenes as a precursor severely limits the applicability of a method. With the advance of 
neural networks (CNNs), generative adversarial networks~\cite{GAN} (GANs) achieve impressive results in image generation, allowing view synthesis without
explicit geometrical structures of objects or scenes.

\noindent\myparagraph{View synthesis via embedded geometry}
Zhou~\etal~\cite{Zhou2016} proposes learning a flow field that maps pixels in 
input images to their corresponding locations in target views to capture
latent geometrical information. \cite{Olszewski2019} learns a volumetric 
representation in a transformable bottleneck layer, which can generate corresponding views for arbitrary transformations. The former explicitly utilizes input (source) image pixels in constructing new views, either fully~\cite{Zhou2016}, or partly with the rest being filled by a completion network~\cite{Park2017,Sun2018}. The latter explicitly applies transformations on the volumetric representation in latent space and generates new views by means of pixel generation networks.

The proposed method takes the best of both worlds. By directly using object geometry the source pixels are mapped to their target positions based on given transformation parameters, hence making the best use of the given information synthesizing new views. \CR{Our approach is fundamentally different from~\cite{Park2017}: we estimate the object point cloud using self-supervised depth predictions and obtain coarse target views from purely geometrical transformations, while~\cite{Park2017} learns mappings from input images and ground truth occluded regions to generate coarse target views using one-hot encoded transformation vectors.}

\noindent\myparagraph{View synthesis directly from image}
Since the introduction of image-to-image translation~\cite{Isola2017},
there is a paradigm shift towards pure image-based approaches~\cite{Tatarchenko2016}.
\cite{Zhu2018} synthesizes bird 
view images from a single frontal view image, while \cite{Regmi2018} generates
cross-views of aerial and street-view images. The networks can be trained
to predict all the views in an orbit from a single-view 
object~\cite{Kicanaoglu2018,Johnston2019}, or generate a view in an iterative 
manner~\cite{Galama2019}. Additional features can be embedded such as 
view-independent intrinsic properties of objects \cite{Xu2019_ICCV}.
In this paper, we employ GANs to generate complete views, which is conditioned
on the geometrical features and the relative poses between source and target
views.
Our approach can be interpreted as a reverse and end-to-end process
of~\cite{Johnston2019}: we estimate objects' arbitrary new views via point clouds
constructed from self-supervised depth maps, while~\cite{Johnston2019} predict objects' fixed orbit views for 3D reconstruction.


\section{Method}

\subsection{Point-cloud based transformations}
\label{subsec:pc_transf}
The core of the proposed novel view synthesis method is to use point clouds for geometrically aware transformations. 
Using the pinhole camera model and known intrinsics  $\bf K$, the point cloud can be reconstructed when the pixel-wise depth map (D) is available.
The camera intrinsics can be obtained by camera calibration, yet for the synthetic data used in our experiments, $\bf K$ is given. 
A pixel on the source image plane $p_s = [u, v, 1]$ (using homogeneous coordinates), corresponds to a point $P_s = [X, Y, Z]$ in the source camera space:
\begin{align}
    D_s \ p_s^\top &= {\bf K} \ P_s^\top & P_s^\top &= {\bf K}^{-1} \ D_s \ p_s^\top
    \label{eq:cam_model}    
\end{align}
Rigid transformations can be obtained by matrix multiplications. The relative transformation to the \emph{target} viewpoint from the \emph{source} camera, is given by
\begin{equation}
    \theta_{s\rightarrow t} = \left[ {\begin{array}{*{20}{c|c}} \bf R & \bf t \\ \hline 0 & 1 \end{array}} \right]
\end{equation}
where ${\bf R}$ denotes the desired rotation matrix and ${\bf t}$ the translation vector.
Points in the target camera view are given by $P_t = \theta_{s\rightarrow t}P_s$.
This can also be regarded as an image-based flow field
$\phi:\mathbb{R}^2 \rightarrow \mathbb{R}^2$ parameterized by
$\theta_{s\rightarrow t}$ (\cf~\cite{Zhou2016,Park2017,Sun2018}).
The flow field $\phi(p_s; \theta_{s\rightarrow t})$ returns the homogeneous coordinates of pixels in the target image for each pixel in the source image:
\begin{equation}
        \phi(p_s; \theta_{s\rightarrow t}) = {\bf K} \ \theta_{s\rightarrow t} \ {\bf K}^{-1} \ D_s p_s^\top 
        \label{eq:flow}
\end{equation}
By observing that $\phi(p_s; \theta_{s\rightarrow t}) = D_t p_t$, the Cartesian pixel coordinates in the target view can be extracted.
The advantage of the flow field interpretation is that it provides a direct mapping between the image planes of the source view and the target view.

\begin{figure*}[t]
    \centering
    \includegraphics[width=.8\textwidth]{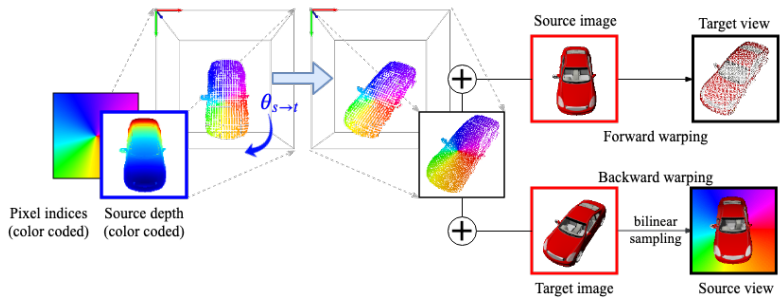}
    \caption{Illustration of the forward and backward warping operation of point clouds.
    The forward warping is used to generate a coarse target view, while the backward warping is used to reconstruct the source view from a target view for self-supervised depth estimation.} 
    \label{fig:forward_backward_warping}
    \vspace{-3mm}
\end{figure*}

\myparagraph{Forward warping}
The flow field is used to generate the target view from the source:
\begin{equation}
    \tilde{I}_t(\phi(p_s;\theta_{s\rightarrow t})) = I_s(p_s).
    \label{eq:forward}
\end{equation}
The resulted image is sparse due to the discrete pixel coordinates and (dis)occluded regions, see Fig.~\ref{fig:forward_backward_warping} (\emph{top-right}).
It is used as input to the image completion network (Sec.~\ref{subsec:completion}).

\myparagraph{Backward warping}
The flow field is used to generate the source view from the target:
\begin{equation}
    \tilde{I}_s(p_s) = I_t(\phi(p_s; \theta_{s\rightarrow t})).
    \label{eq:backward}
\end{equation}
The process assigns a value to every pixel $(u,v)$ in $\tilde{I}_s$ resulting in a dense image, as illustrated in Fig.~\ref{fig:forward_backward_warping} (\emph{bottom-right}).
The generated source view may contain artifacts due to (dis)occlusion in the target view.
To sample $\phi(p_s;\theta_{s\rightarrow t})$ from $I_t$, a differentiable bi-linear sampling layer~\cite{Jaderberg2015} is used.
The generated source view is used for self-supervised monocular depth prediction (Sec.~\ref{subsec:mono_depth}). 

\subsection{Novel view synthesis}
\label{subsec:completion}

The point-cloud-based forward warping relocates the visible pixels of the object in the source view to their corresponding positions in the target view. For novel view synthesis, however, two more steps are required:
(1) obtaining the target coarse view by discarding occluded pixels, and (2) filling in the pixels that are not seen in the source view. 

\myparagraph{Coarse view construction}
The goal is to remove the pixels which are seen in the source view yet should not be visible in the target view, due to occlusion.
To this end, pixels that have surface normals (after transformation) pointing away from the viewing direction are removed, similarly to~\cite{Park2017}. 
Surface normals are obtained from normalized depth gradients.

\begin{figure}[t]
    \centering
    \includegraphics[width=.99\textwidth]{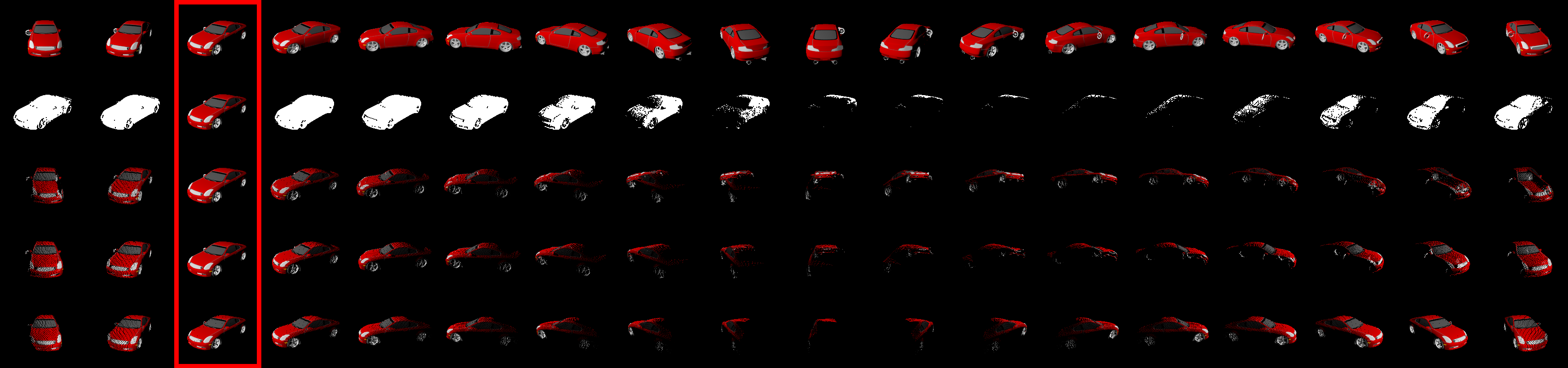}
    \caption{Image coarse views for different target viewpoints. 
    The input image is depicted in the third column (red box). 
    From top to bottom: (1) target views, (2) source region visible in each target viewpoint, coarse view (3) naive (4) with occlusion removal, and (5) with occlusion removal and symmetry. }
    \label{fig:allview_with_mask}
\end{figure}

An illustration of the coarse view construction is shown in Fig.~\ref{fig:allview_with_mask} for different target views.
The first row depicts the target views, the second row indicates the visible parts from the input image (third column).
The third and fourth row show the coarse view with and without occlusion removal (or backface culling).
Finally, the fifth row shows an enhanced version of the coarse view, where the object is assumed to be left-right symmetric~\cite{Park2017}.
The proposed method directly identifies and removes occlusion pixels from the input view using \emph{estimated} depth, which contrasts to~\cite{Park2017}, where ground truth visibility mask are required for each target view to train a visibility prediction network.

\myparagraph{View completion}
The obtained coarse view is already in the target viewpoint, but it remains sparse. To synthesize the final dense image, an image completion network is used.

The completion network uses the hour-glass architecture~\cite{Newell2016}.
Following~\cite{Park2017}, we concatenate the depth bottleneck
features and embedded transformation to the completion network
bottleneck. By conditioning the completion network on the input features and the desired transformation $\theta_{s\rightarrow t}$,
the network can fix artifacts and errors due to estimated depth and cope better with extreme pose transformations, \ie when coarse view image is near
empty (\eg columns 9-11 in Fig.~\ref{fig:allview_with_mask}). 

The image completion network is trained in a GAN-manner by using a generator $\mathcal{G}$, a discriminator $\mathcal{D}$, an input image $I_s$ and a target image $I_t$.
The combination of losses that are used is given by 
\begin{align}
    \label{eq:lsgan}
    \mathcal{L}_{\mathcal{D}} &= \left(\mathcal{D}(I_s) - 1\right)^2 + \mathcal{D}(\mathcal{G}(I_s)))^2, & &\text{LS-GAN Discriminator loss}\\
    \label{eq:genloss}
    \mathcal{L}_{\mathcal{G}} &= \left[1 - \text{SSIM}(I_t, \mathcal{G}(I_s))\right] + \left\|I_t - \mathcal{G}(I_s)\right\|_1, & &\text{Generator loss}\\
    \label{eq:percloss}
    \mathcal{L}_{Perc} &= \big\| \mathcal{F}^{\mathcal{D}}_{I_t} - \mathcal{F}^{\mathcal{D}}_{\mathcal{G}(I_s)}\big\|_2 + \big\| \mathcal{F}^{\text{VGG}}_{I_t} - \mathcal{F}^{\textrm{VGG}}_{\mathcal{G}(I_s)}\big\|_2, & &\textrm{Perceptual loss}
\end{align}
where the perceptual loss uses $\mathcal{F}^{\mathcal{D}}$ ($\mathcal{F}^{\text{VGG}}$) to denote features extracted from image $I_t$ and $\mathcal{G}(I_s)$ from the discriminator network and pre-trained VGG network respectively, \cf~\cite{Perceptual}. \CR{SSIM denotes the structural similarity index measure, see Sec.~\ref{sec:exp}.} 
The total loss is given by:
\begin{equation}
    \mathcal{L}_c = w_1 \mathcal{L}_{\mathcal{D}}  + w_2 \mathcal{L}_{\mathcal{G}} + w_3 \mathcal{L}_{Perc},
\end{equation}
where $w$ denotes the weighting of the losses ($w_1=1, w_2=100$, and $w_3=100$, \cf~\cite{Park2017}).

\subsection{Self-supervised Monocular Depth estimation}
\label{subsec:mono_depth}
The discussion so far has assumed that pixel-wise depth maps are available. 
In this section, the method used to estimate depth from a single $RGB$ image is detailed.
In order to make the minimum assumption about the training data, self-supervised methods are considered, which do not require ground-truth depth~\cite{Garg2016,Zhou2017,Godard2017,Godard2019,Johnston2019}.

For the depth prediction an encoder-decoder network with bottleneck architecture is used, similar to~\cite{Godard2019}. The network is optimised using a set of (reconstruction) losses between the source image $I_s$ and its synthesized version $\tilde{I}_s$, using the backward warping, Eq.~ \eqref{eq:backward}, from a second (target) image $I_t$ and the predicted depth map.
The underlying rationale is that a more realistic depth map will have a lower reconstruction loss. 

The losses are given by:
\begin{align}
    \label{eq:peloss}
    \mathcal{L}_{p}\left(I_s, \tilde{I}_s\right) &= \frac{\alpha}{2}\left[1-\text{SSIM}\left(I_s, \tilde{I}_s\right)\right] + (1-\alpha)\left\|I_s - \tilde{I}_s\right\|_1, & & \text{Photometric loss} \\
    \mathcal{L}_{s}(d) &=\left|\partial_x d\right|e^{-\left|\partial_x I_s\right|} + \left|\partial_y  d \right|e^{-\left|\partial_y I_s\right|}, & &\text{Smoothness loss~\cite{Godard2017}}\\
    \mathcal{L}_{d}\left(I_s, \tilde{I}_s, d\right) &= \mu \mathcal{L}_{p}\left(I_s, \tilde{I}_s\right) + w_d \mathcal{L}_{s},    & & \text{Total loss}\label{eq:smoothloss}
\end{align}
\noindent where $\alpha=0.85$, and $d = \frac{\overline{D}}{D}$ is the mean-normalized inverse depth, $w_d=10^{-3}$, and $\mu$ is an indicator function which equals 1 iff the photometric loss $\mathcal{L}_p(I_s,\tilde{I}_s)<\mathcal{L}_p(I_s,I_t)$, see~\cite{Godard2019} for more details. The smoothness loss encourages nearby pixels to have similar depths, while the artifacts due to (dis)occlusion are excluded by the per-pixel minimum-projection mechanism.

\section{Experiments}
\label{sec:exp}
In this section, the proposed method is analysed on the 3D ShapeNet benchmark including an ablation to study the effects of the different components and a state-of-the-art comparison.

\myparagraph{Dataset}
We use the object-centered car and chair images rendered from the 3D ShapeNet 
models~\cite{shapenet} using the same render engine\footnote{The specific render engine and setup is to guarantee fair comparison with reported methods as none of the authors-provided weights
perform at the similar level on images rendered with different rendering setups.}
and set up as in~\cite{Zhou2016,Park2017,Sun2018,Olszewski2019}.
Specifically, there are
7497 car and 698 chair models with high-quality textures, split by 80\%/20\% for training and test. The images are rendered
at 18 azimuth angles (in $[0, 340]$, $20^{\circ}$-separation) and 3
elevation angles ($0^{\circ}, 10^{\circ}, 20^{\circ}$).
Input and output images are of size $256\times256$.

\myparagraph{Metrics} We evaluate the generated images using the
standard $L_1$ pixel-wise error (normalized to the ranged $[0, 1]$, lower is 
better) and the structural similarity index measure (SSIM)~\cite{ssim} (value 
range of $[-1, 1]$, higher is better). $L_1$ indicates the proximity of pixel 
values between a completed image and the target, while SSIM measures the 
perceived quality and structural similarity between the images.

\myparagraph{Baseline} We compare the results of our method with the following
state-of-the-art methods: AFN~\cite{Zhou2016}, TVSN~\cite{Park2017}, 
M2NV~\cite{Sun2018}, and TBN~\cite{Olszewski2019}.

\subsection{Initial experiments}
\myparagraph{Comparison to image-based completion}
In this section, we compare the intermediate views generated by
the forward warping using  estimated point clouds and those by image-based flow field 
prediction by DOAFN~\cite{Park2017} and M2NV~\cite{Sun2018}.
For this experiment, the coarse view after occlusion removal and left-right symmetric enhancements are used. 
The image completion network is the basis variant, using DCGAN, without bottleneck inter-connections.
The results are shown in Table~\ref{tab:coarse_view}.
The transformation of estimated point clouds
provides coarse views which are closer to the target view, and these help to obtain a higher quality of completed views.

\begin{table}[t]
    \caption{
    Quality of coarse and completed view (\emph{a}) and ablation study (\emph{b}).
    The proposed transformation within estimated point clouds generates better
    coarse images over image-based predicted flow fields. Subsequently,
    the completed view quality is improved. The ablation study shows the best
    results are obtained by using LSGAN (LS), perceptual loss (PL), symmetry (Sym), bottleneck inter-connection (IC), and SSIM loss (SL).}
    \label{tab:joint_coarse_and_ablation}
    \subtable[][Coarse vs Completed]{
        \centering
        \resizebox{.6\textwidth}{!}{
            \begin{tabular}{@{}llllll@{}}
            \toprule
            & \multicolumn{2}{c}{Coarse view} & \phantom{a}& \multicolumn{2}{c}{Completed view} \\
            \cmidrule{2-3} \cmidrule{5-6}
            & L1 $(\downarrow)$ & SSIM $(\uparrow)$ & & L1 $(\downarrow)$ & SSIM $(\uparrow)$ \\
            \midrule
            DOAFN \cite{Park2017} & .220 & .876 && .121 & .910 \\
            M2NV \cite{Sun2018} & .226 & .879 && .154 & .906 \\
            Ours & \bf .203 & \bf .882 && \bf .118 & \bf .924  \\
            \bottomrule
            \end{tabular}
        }
        \label{tab:coarse_view}
    }
    \subtable[][Ablation study]{
        \centering
        \resizebox{.39\textwidth}{!}{
            \begin{tabular}{@{}ccccccc@{}}
                \toprule
                LS & PL & Sym & IC & SL & L1 $(\downarrow)$ & SSIM $(\uparrow)$ \\
                \midrule
                & \checkmark & \checkmark & & & .118 & .924 \\
                \checkmark & \checkmark & \checkmark & &  & .101 & .939 \\
                \checkmark & & \checkmark & & & .103 & .939 \\
                \checkmark & \checkmark & & & & .107 & .933 \\
                \checkmark & \checkmark & \checkmark & \checkmark &  & \bf .097 & .939 \\
                \checkmark & \checkmark & \checkmark & \checkmark & \checkmark & \bf .097 & \bf .942 \\
                \bottomrule
            \end{tabular}
        }
        \label{tab:ablation_study}
    }
\end{table}

\paragraph{Ablation Study}
We analyze the effects of the different component of the proposed pipeline. The
results are shown in Table~\ref{tab:ablation_study}. The use of the LSGAN loss shows a relative large improvement over the traditional DCGAN. The drop of performance by removing symmetry assumption shows the importance of prior knowledge on target objects, which is intuitive.
The inter-connection from the depth network and the embedded transformation to the completion network allow the model to not rely solely on intermediate views. This is important for overcoming errors and artifacts which occur in the coarse images (due to inevitable uncertainties in depth prediction) and generate in general higher quality images. The SSIM loss, first employed by~\cite{Olszewski2019}, shows improvement in SSIM metric, which is intuitive as training objectives are closer to evaluation metrics.

\subsection{Comparison to State-of-the-Art}

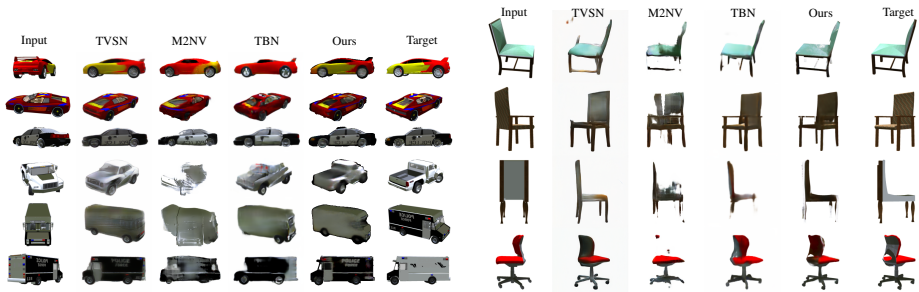
\begin{figure}[t]
    \centering
    \resizebox{.48\textwidth}{!}{
    \begin{tikzpicture}
        \node {\includegraphics[width=\textwidth]{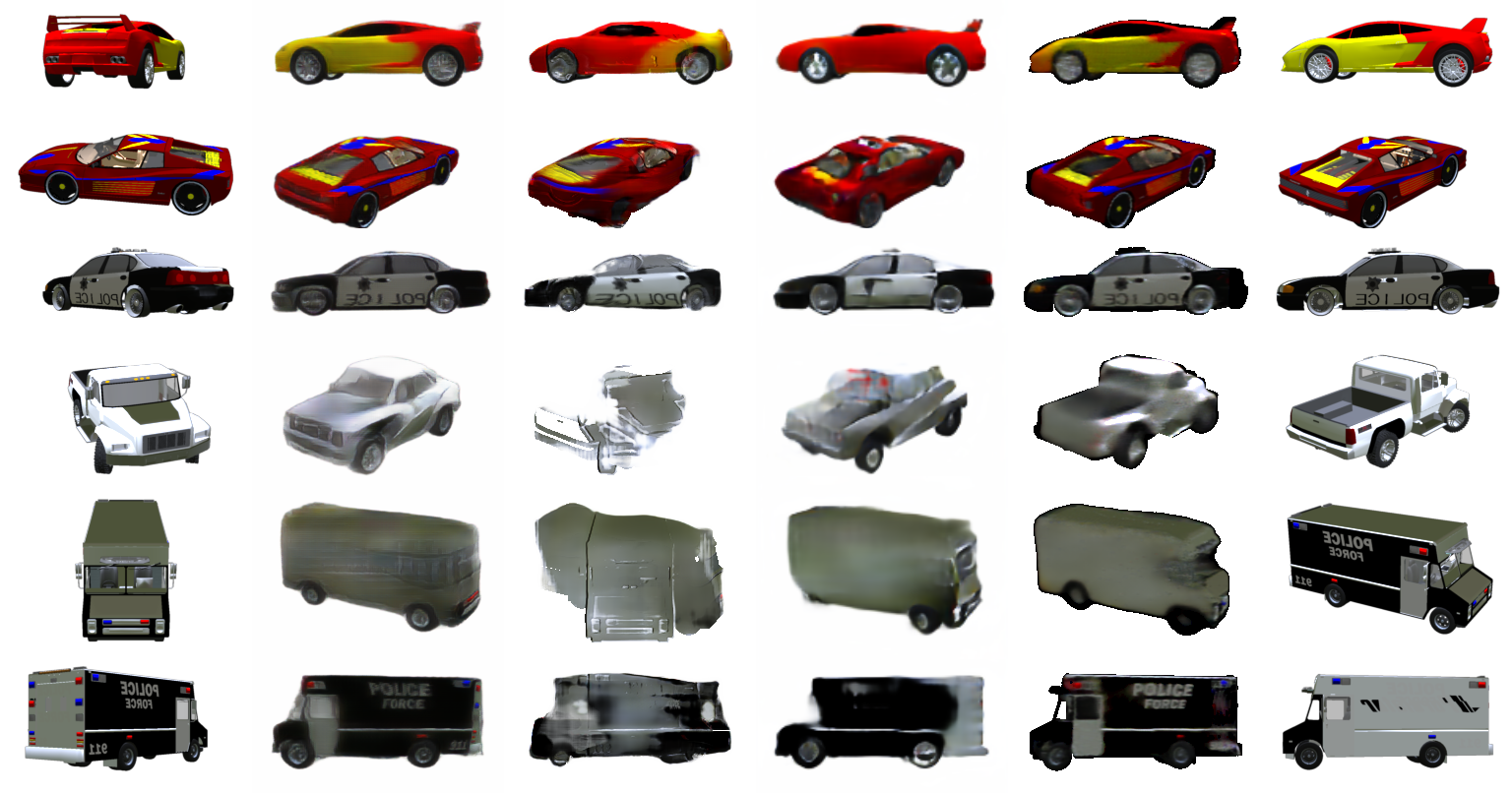}};
        \node at (-5.5,3.7) {Input};
        \node at (-3.3,3.7) {TVSN};
        \node at (-1.1,3.7) {M2NV};
        \node at ( 1.1,3.7) {TBN};
        \node at ( 3.3,3.7) {Ours};
        \node at ( 5.5,3.7) {Target};
    \end{tikzpicture}
    }
    \resizebox{.48\textwidth}{!}{
    \begin{tikzpicture}
        \node {\includegraphics[width=\textwidth]{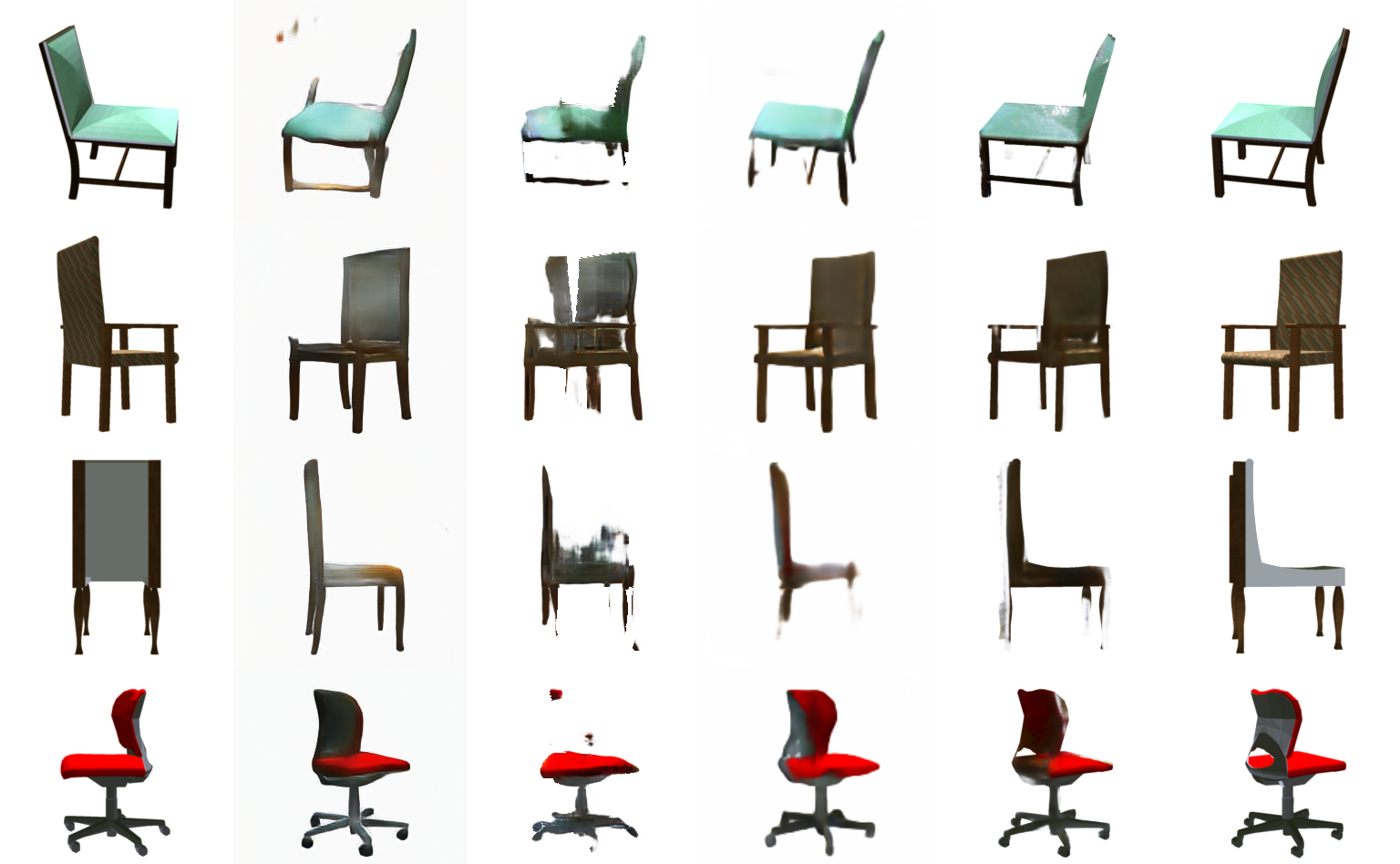}};
        \node at (-5.4,3.9) {Input};
        \node at (-3.3,3.9) {TVSN};
        \node at (-1.1,3.9) {M2NV};
        \node at ( 1.1,3.9) {TBN};
        \node at ( 3.3,3.9) {Ours};
        \node at ( 5.5,3.9) {Target};
    \end{tikzpicture}
    }
    \caption{
    Qualitative comparisons of synthesized cars (\emph{left}) and chairs (\emph{right}), given  
    a single input image (\emph{first column}) and a given target view (\emph{last column}). The last two rows show a more challenging examples. The proposed method captures better the geometry of the object and the fine (texture) details.
    More examples are provided in the supplementary materials.}
    \label{fig:sota_qualitative}
    \vspace{-3mm}
\end{figure}

In this section, the proposed method is compared with state-of-the-art methods.
The quantitative results are shown in Table~\ref{tab:SOTA}.
The proposed method performs consistently performs (slightly) better on both evaluation metrics for both types of objects.
Quantitative results are shown in 
Fig.~\ref{fig:sota_qualitative} where challenging cases are shown in the last 2 rows. Notice the better ability in retaining objects' textures (such as color patterns and texts on cars) of methods that explicitly use input pixel values in generating new views to that of TBN.
The results of cars are constantly higher than that of
chairs due to the intricate structures of chairs. However, by having access to object geometry, geometrical assumptions such as symmetry and occlusion can be applied directly to intermediate views (instead of having to learn from annotated data~\cf~\cite{Park2017}), which creates better views for near-to symmetry targets.
High-quality qualitative results and more analyses can be found in the supplementary materials.

Table~\ref{tab:SOTA} also shows the evaluation when target viewpoints are from 
different elevation angles. Methods such AFN, TVSN, and M2NV encode 
transformation as one-hot vectors and thus, are limited to operate within a 
pre-defined set of transformations (18 azimuth angles, same elevation).
This is not the case for our method and TBN which apply direct transformation.
We use the same azimuth angles as in the standard test set while randomly sample
new elevation angles for input images in $(0^\circ, 10^\circ, 20^\circ)$.
The results are shown with networks trained with the regular fixed-elevation 
settings.
The new transformations produces different statistics from what the networks have
been trained, resulting in a performance drop for both methods. Nevertheless,
the proposed method can still maintain high quality image synthesis.

\begin{table*}[t]
    \centering
    \resizebox{.6\textwidth}{!}{
    \begin{tabular}{@{}llllll@{}}
        \toprule
        \phantom{abc}&\multirow{2}{*}{Methods} & \multicolumn{2}{c}{cars} &  \multicolumn{2}{c}{chairs}  \\
        \cmidrule(r){3-4}\cmidrule(l){5-6}
        &&  L1 $(\downarrow)$ & SSIM $(\uparrow)$ & L1 $(\downarrow)$ & SSIM $(\uparrow)$ \\
        \midrule
        \multicolumn{2}{l}{\textit{Same elevation}} \\
        &AFN~\cite{Zhou2016} & .148 & .877 & .229 & .871\\
        &TVSN~\cite{Park2017} & .119  & .913 & .202 & .889 \\ 
        &M2VN~\cite{Sun2018} & .098 & .923 & .181 & .895 \\
        &TBN~\cite{Olszewski2019} &  .091 & .927 & .178 & .895\\
        &ours & \bf .096  & \bf .945 & \bf .175 & \bf .914  \\
        \midrule
        \multicolumn{2}{l}{\textit{Cross-elevation}} \\
        &TBN &  .199  & .910 & .215 & .902 \\
        &ours & \bf .122  & \bf .934 & \bf .207 & \bf .905  \\
        \bottomrule
    \end{tabular}
    }
    \caption{Quantitative comparison with state-of-the-art methods on novel view synthesis: our method consistently performs (slightly) better than the other methods for both categories where target views have the same or different elevation angles with input views.}
    \label{tab:SOTA}
    \vspace{-3mm}
\end{table*}
    
\subsection{Multi-View Synthesis and Point Cloud Reconstruction}

\myparagraph{Multi-view inputs}
The proposed method can be naturally extended to use multi-view inputs as follows:
for each image depth is predicted independently and combined into a single point cloud.
The resulting coarse target image will be denser when more images are used, and is passed through the image completion network.

In this experiment, the model trained for single-view prediction is used and evaluated using multiple (1 to 8) input images. 
The results in Table~\ref{tab:multiview} show that
the quality of the coarse view increases, as expected,  when more input images are used and hence the point clouds are denser. 
Surprisingly, however, the image completion network only marginally improves, indicating that the coarse view contains enough information for the image completion network to synthesis a high quality target image.

\begin{wraptable}{r}{0.5\textwidth}
    \centering
    \resizebox{.5\textwidth}{!}{
    \begin{tabular}{ccccc}
    \toprule
        \multirow{2.5}{*}{\shortstack[c]{No.\\ views}} & \multicolumn{2}{c}{Coarse} & \multicolumn{2}{c}{Final}\\
        \cmidrule(r){2-3}\cmidrule(l){4-5}
        & L1 ($\downarrow$) & SSIM ($\uparrow$) & L1 ($\downarrow$) & SSIM ($\uparrow$)\\
        \midrule
        1 & .203 & .882 & .090 & .945\\
        2 & .188 & .888 & .089 & .945 \\
        4 & .152 & .906 & .085 & .946 \\
        8 & \bf .111 & \bf .907 & \bf .084 & \bf .947 \\
        \bottomrule
    \end{tabular}
    }
    \caption{Performance by extending single-view-trained networks for multi-view inputs.
    }
    \label{tab:multiview}
\end{wraptable}

\begin{figure}[t]
    \centering
    \includegraphics[width=\textwidth]{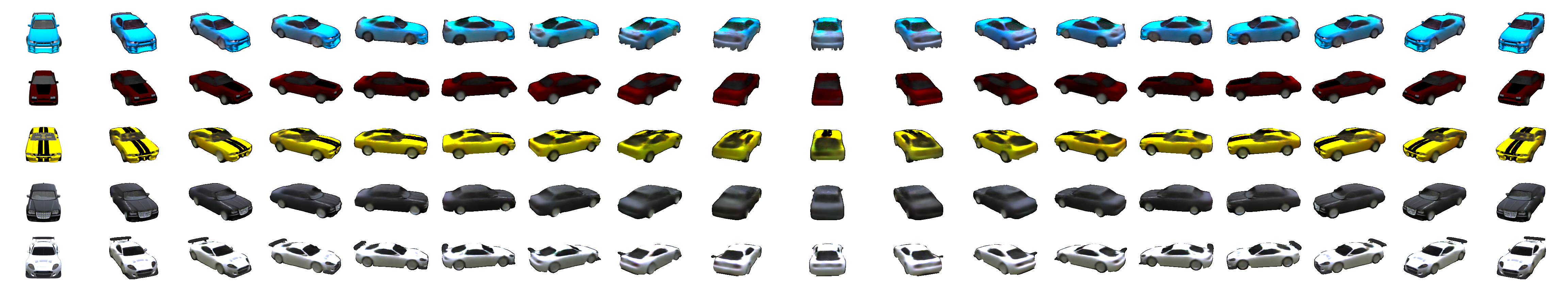} \\
    \includegraphics[width=.18\textwidth]{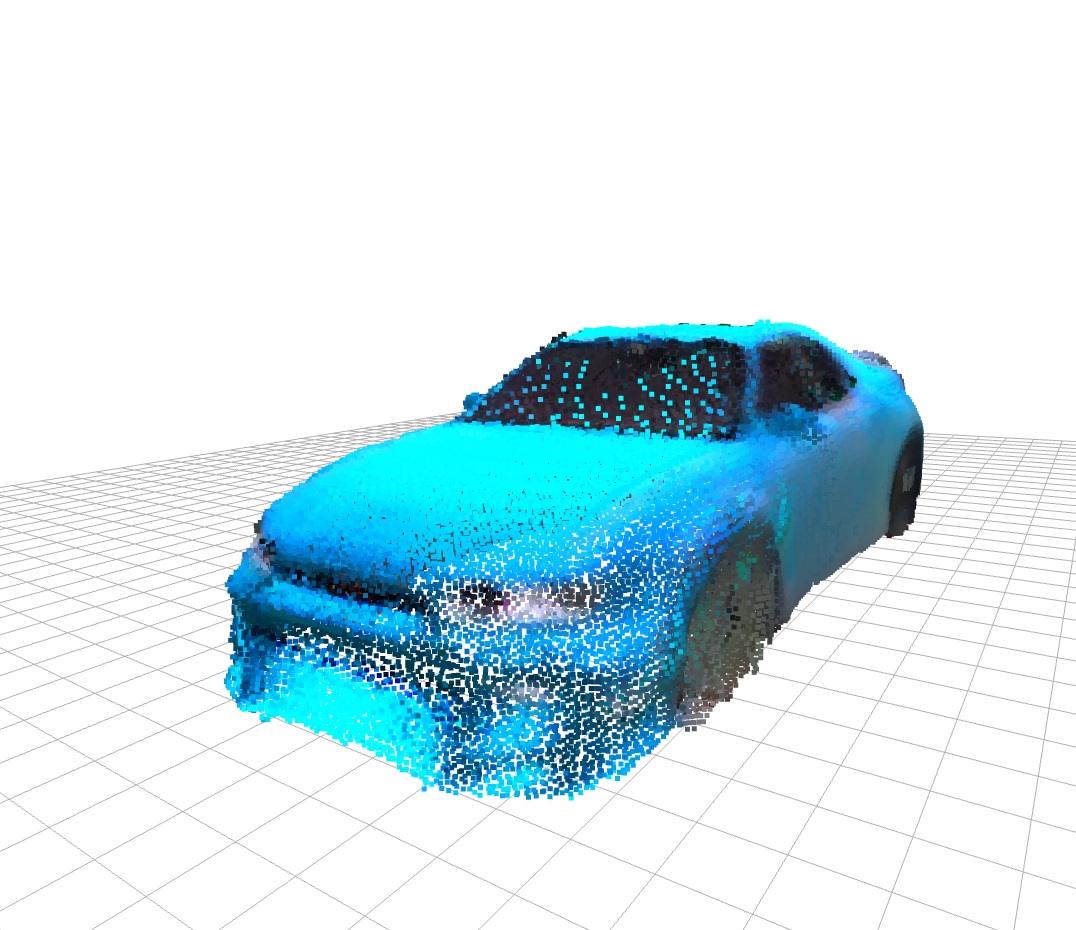}
    \includegraphics[width=.18\textwidth]{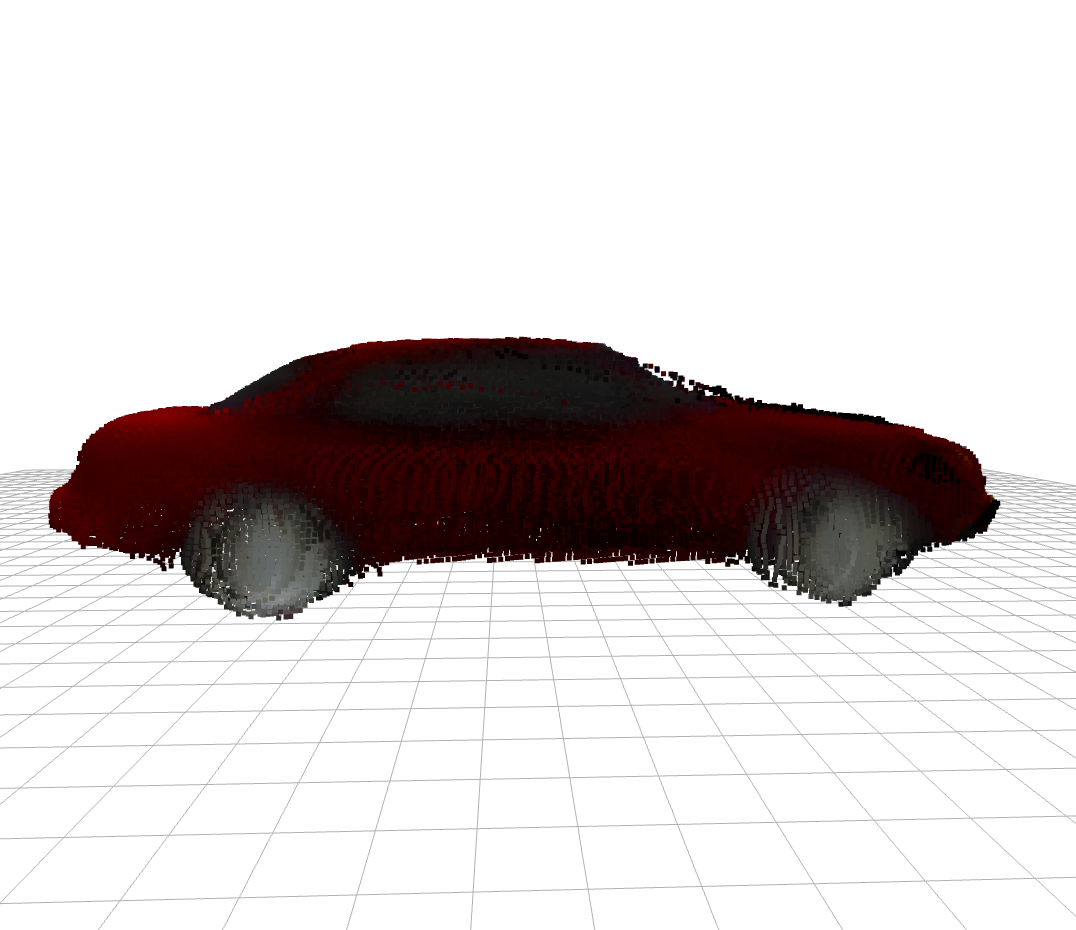}
    \includegraphics[width=.18\textwidth]{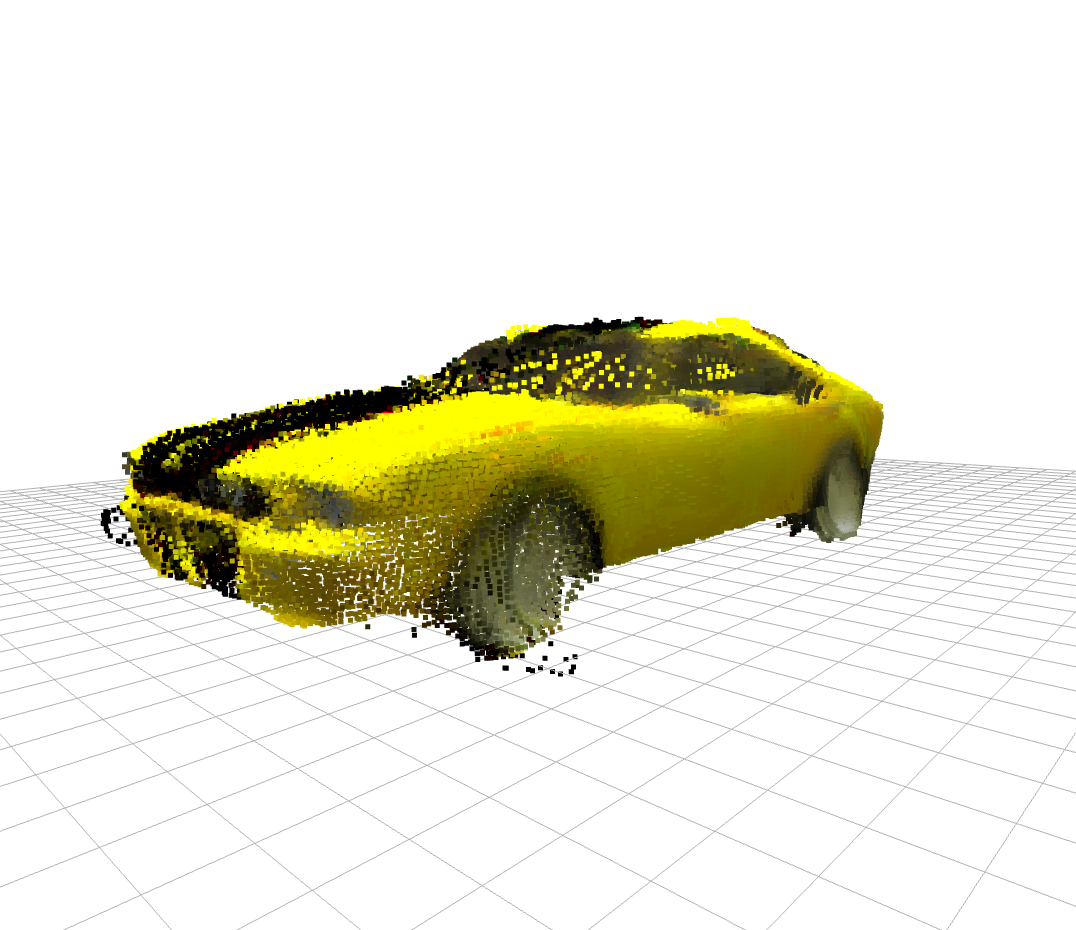}
    \includegraphics[width=.18\textwidth]{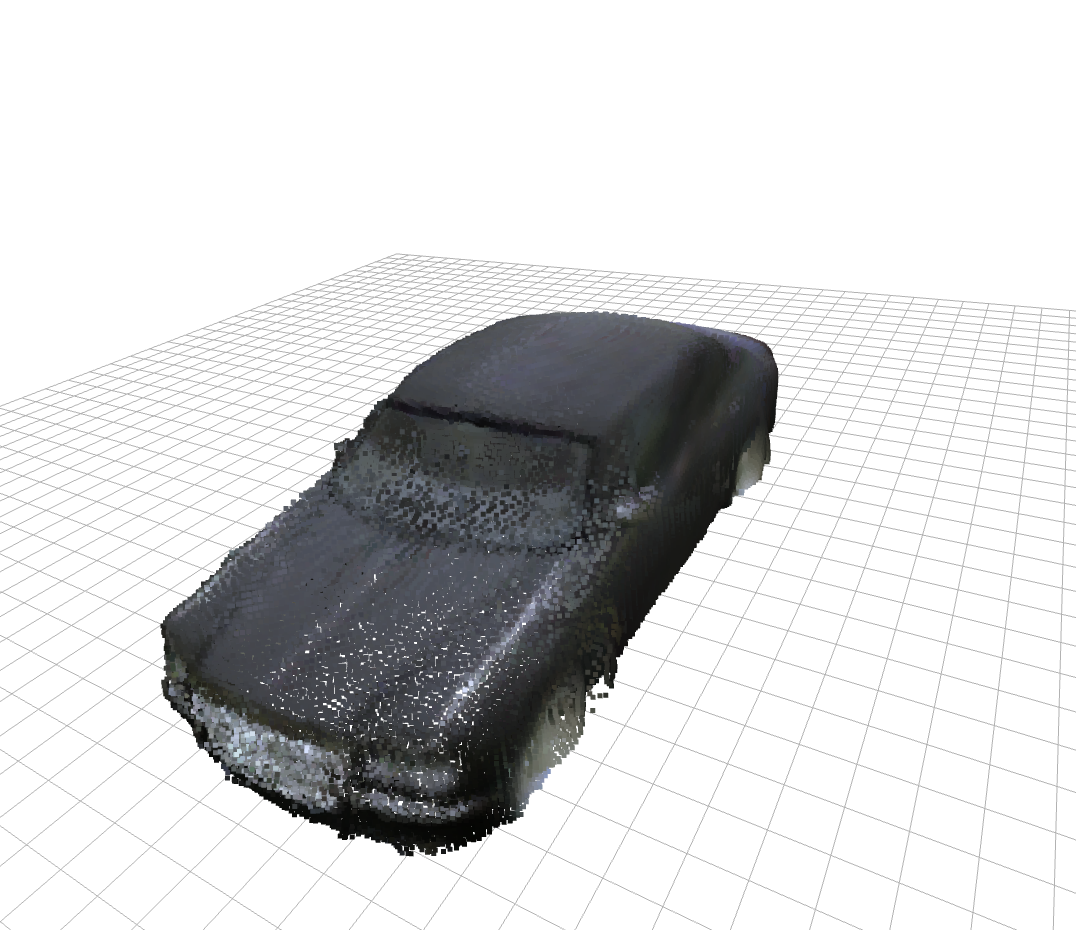}
    \includegraphics[width=.18\textwidth]{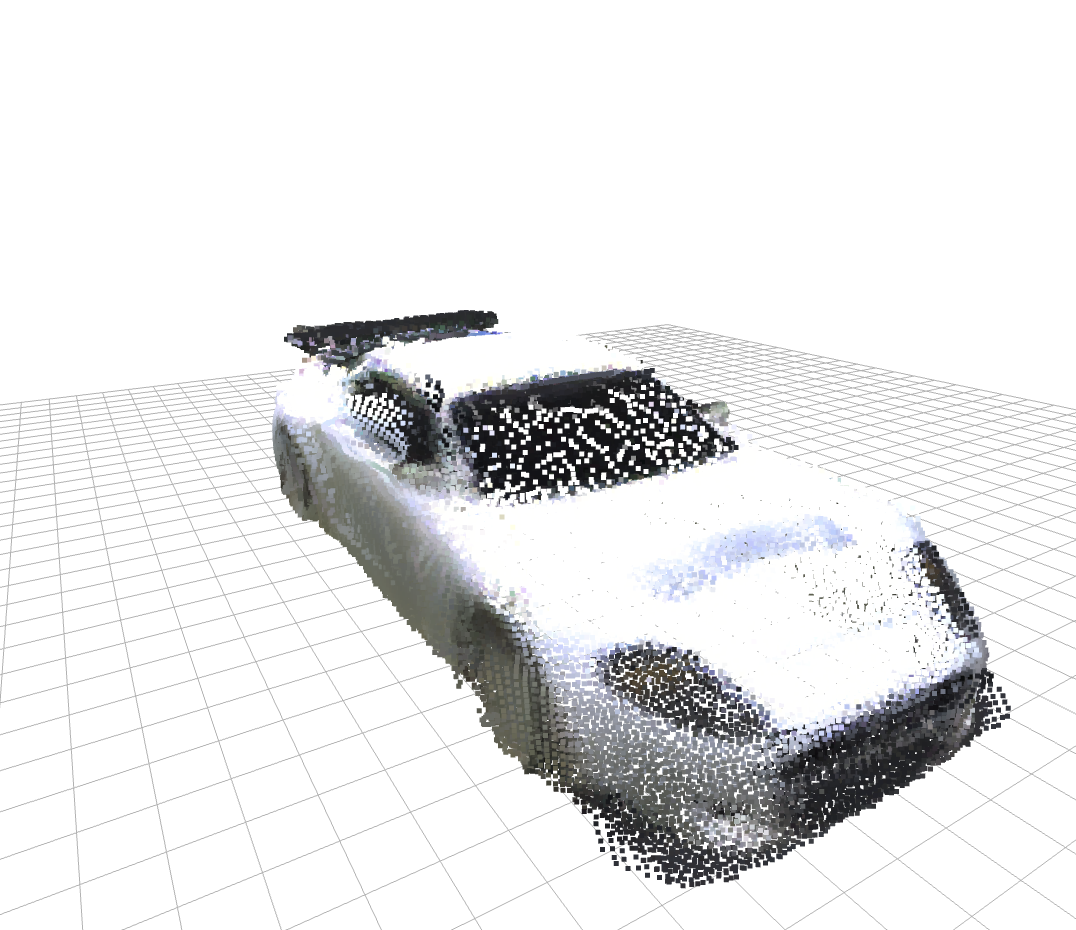}
    \caption{\textit{Top} $360^\circ$-views generation from single input images (first column), \textit{bottom} point cloud reconstruction using estimated depths of each generated views. Depth estimation trained on real images can perform well on synthesized ones.
    }
    \label{fig:pc_construction}
\end{figure}

\paragraph{Point cloud reconstruction}
In this final experiment, the aim is to reconstruct a full dense point cloud from a single image, using the models trained for novel view synthesis.
In order to do so, $360^\circ$-views are generated from a single view of an object, see Fig.~\ref{fig:pc_construction} (top).
Each of these views are fed to the depth estimation network and the obtain estimated depth is used to generate a partial point cloud. 
These point clouds are stitched together, using corresponding 
transformations, resulting in a high quality dense point cloud, as shown 
in Fig.~\ref{fig:pc_construction} (bottom).

\subsection{Results on real-world imagery}

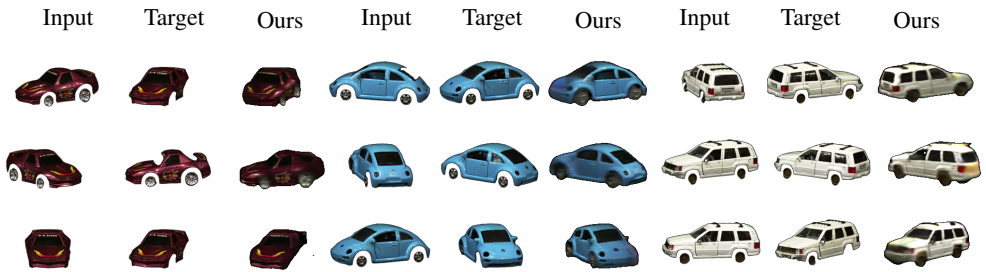
\begin{figure}
    \centering
    \begin{tikzpicture}
        \node {\includegraphics[width=\linewidth]{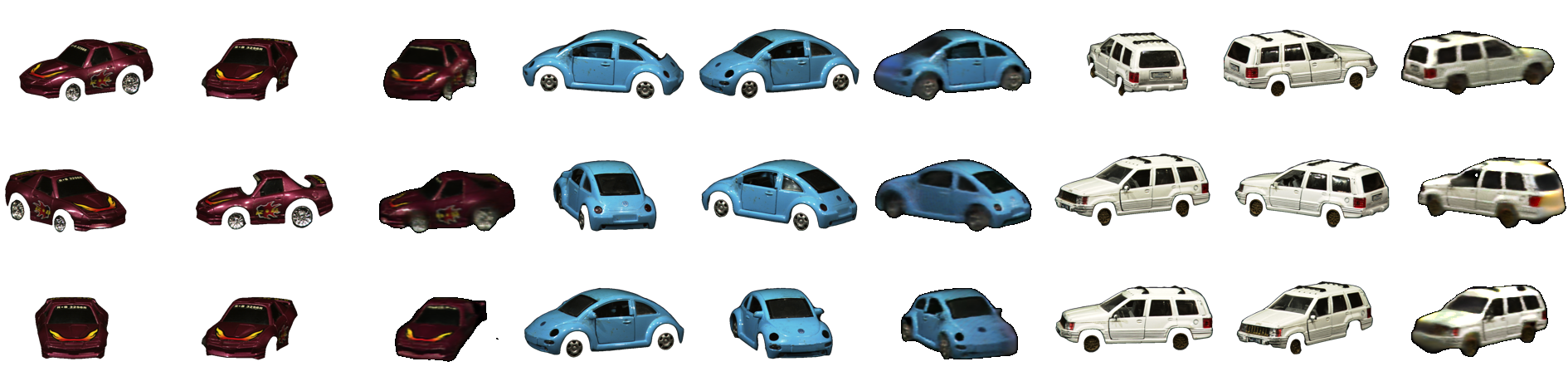}};
        \node at (-5.6,1.9) {\small Input};
        \node at (-4.2,1.9) {\small Target};
        \node at (-2.8,1.9) {\small Ours};
        \node at (-1.4,1.9) {\small Input};
        \node at ( 0.0,1.9) {\small Target};
        \node at ( 1.4,1.9) {\small Ours};
        \node at ( 2.8,1.9) {\small Input};
        \node at ( 4.2,1.9) {\small Target};
        \node at ( 5.6,1.9) {\small Ours};
    \end{tikzpicture}
    \caption{Quantitative results on real-imagery ALOI~\cite{geusebroek2005} dataset. The inputs and targets are shown with provided object masks, while the synthesized images are with predicted masks. The completion network does not need to be finetuned, yet provide competent results.}
    \label{fig:aloi}
\end{figure}

We apply the trained car model to the car images of the real-imagery ALOI 
dataset~\cite{geusebroek2005}, consisting of 100 objects, captured at 72 viewing
angles. We use 4 cars for fine-tuning only the depth network, which requires no 
ground truths, while the image-completion network is left untouched.
The quantivative resuls on the remaining 3 cars are shown in Fig.~\ref{fig:aloi}.

\section{Conclusion}
In this paper partial point clouds are estimated from a single image, by a self-supervised depth prediction network and used to obtain a coarse image in the target view.
The final image is produced by an image completion network which uses the coarse image as input. 
Experimentally the proposed method outperforms any of the current SOTA methods on the ShapeNet Benchmark on novel view synthesis. 
Qualitative results show high quality and dense point clouds, obtained from a single image, by synthesizing and combining $360^\circ$ views.
Based on these results, we conclude that point clouds are a suitable, geometry aware representation for true novel view synthesis.

\bibliography{UvA-bmvc20}

\section{Supplementary materials}

\begin{figure}
    \centering
        \begin{tikzpicture}
        \node{\includegraphics[width=\textwidth]{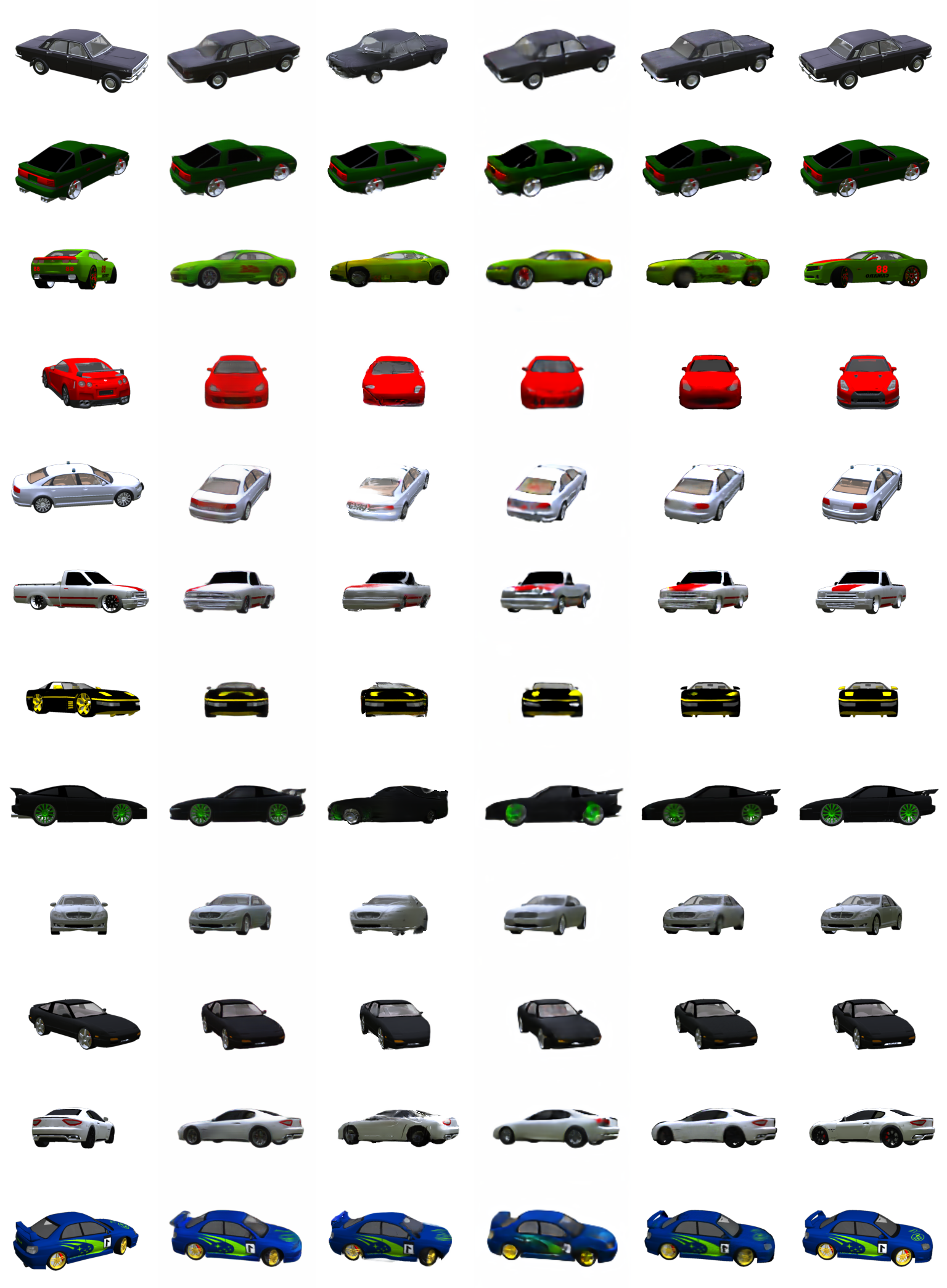}};
        \node at (-5.5,9.0) {Input};
        \node at (-3.3,9.0) {TVSN};
        \node at (-1.1,9.0) {M2NV};
        \node at ( 1.1,9.0) {TBN};
        \node at ( 3.3,9.0) {Ours};
        \node at ( 5.5,9.0) {Target};
    \end{tikzpicture}
    \caption{Qualitative examples of cars generated by our point-cloud-based method and other related work. Our method retains better the objects' geometrical structure and detailed textures, \eg row 3 and 5. }
    \label{fig:sup_car1}
\end{figure}
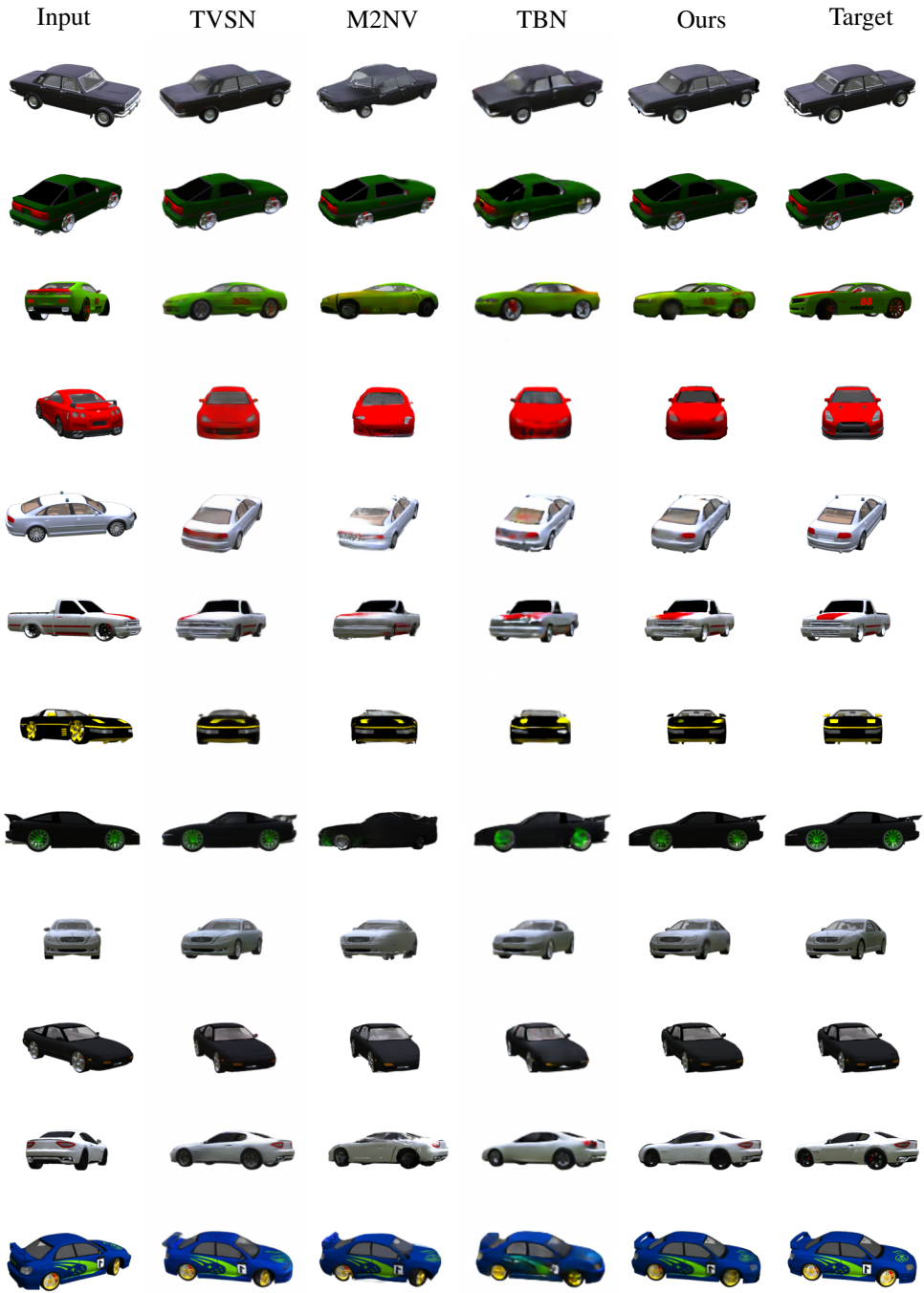

\begin{figure}
    \centering
        \begin{tikzpicture}
        \node{\includegraphics[width=\textwidth]{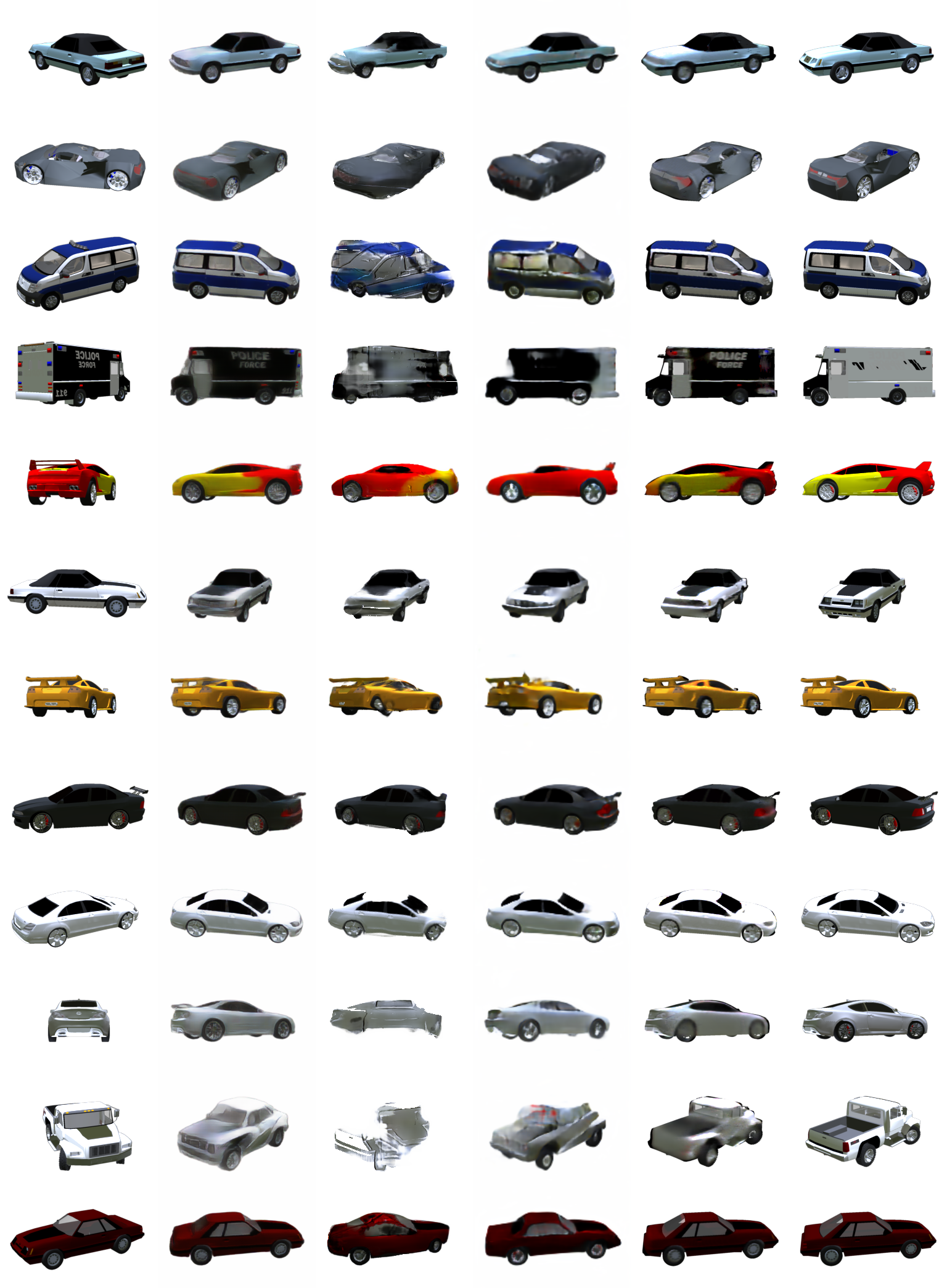}};
        \node at (-5.5,9.0) {Input};
        \node at (-3.3,9.0) {TVSN};
        \node at (-1.1,9.0) {M2NV};
        \node at ( 1.1,9.0) {TBN};
        \node at ( 3.3,9.0) {Ours};
        \node at ( 5.5,9.0) {Target};
    \end{tikzpicture}
    \caption{Qualitative examples of cars generated by our point-cloud-based method and other related work. Our method retains better the objects' geometrical structure and detailed textures, \eg row 5 and 11. }
    \label{fig:sup_car2}
\end{figure}

\begin{figure}
    \centering
        \begin{tikzpicture}
        \node{\includegraphics[width=\textwidth]{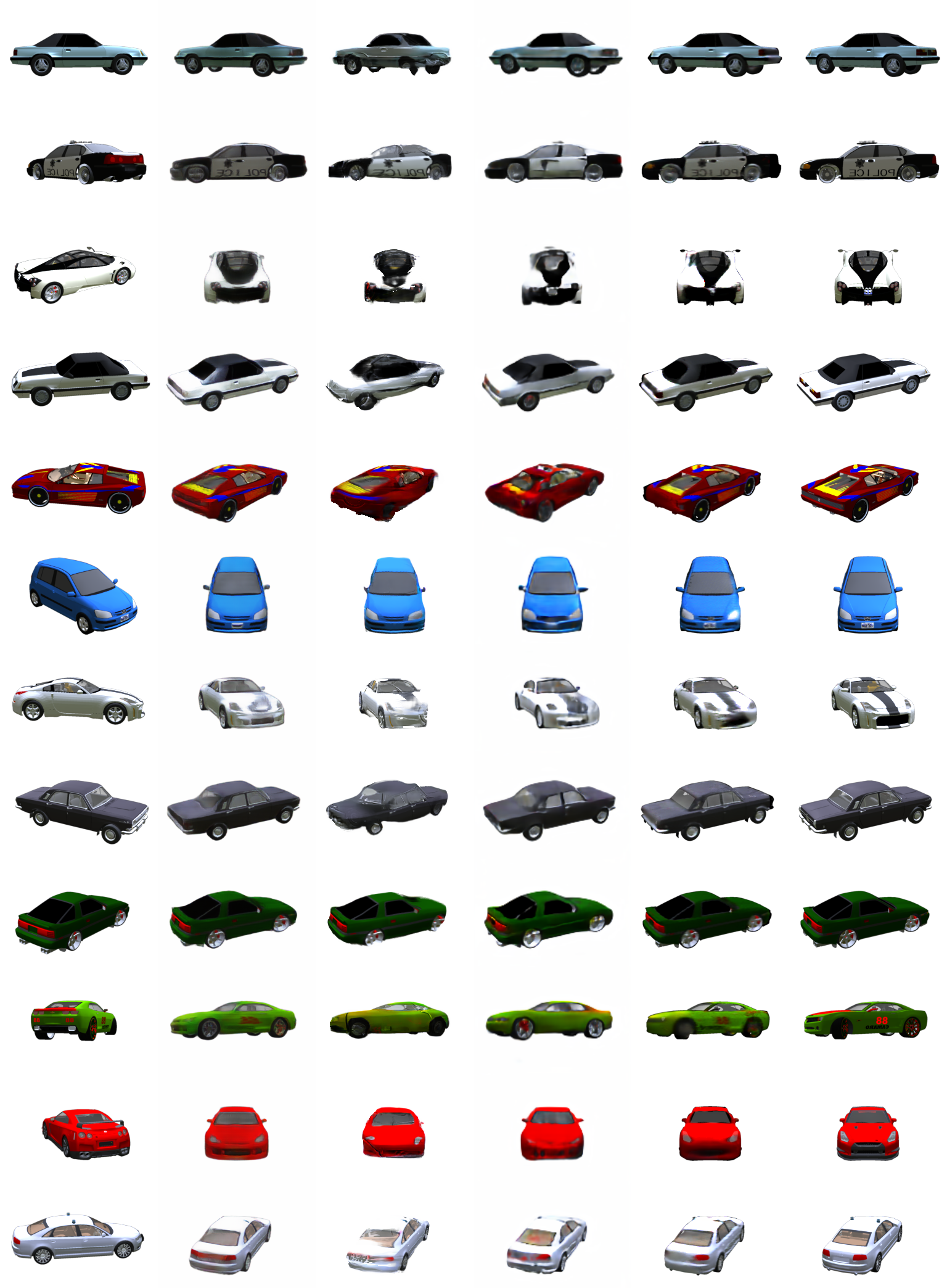}};
        \node at (-5.5,9.0) {Input};
        \node at (-3.3,9.0) {TVSN};
        \node at (-1.1,9.0) {M2NV};
        \node at ( 1.1,9.0) {TBN};
        \node at ( 3.3,9.0) {Ours};
        \node at ( 5.5,9.0) {Target};
    \end{tikzpicture}
    \caption{Qualitative examples of cars generated by our point-cloud-based method and other related work. Our method retains better the objects' geometrical structure and detailed textures, \eg row 2 and 7. }
    \label{fig:sup_car3}
\end{figure}

\begin{figure}
    \centering
        \begin{tikzpicture}
        \node{\includegraphics[width=\textwidth]{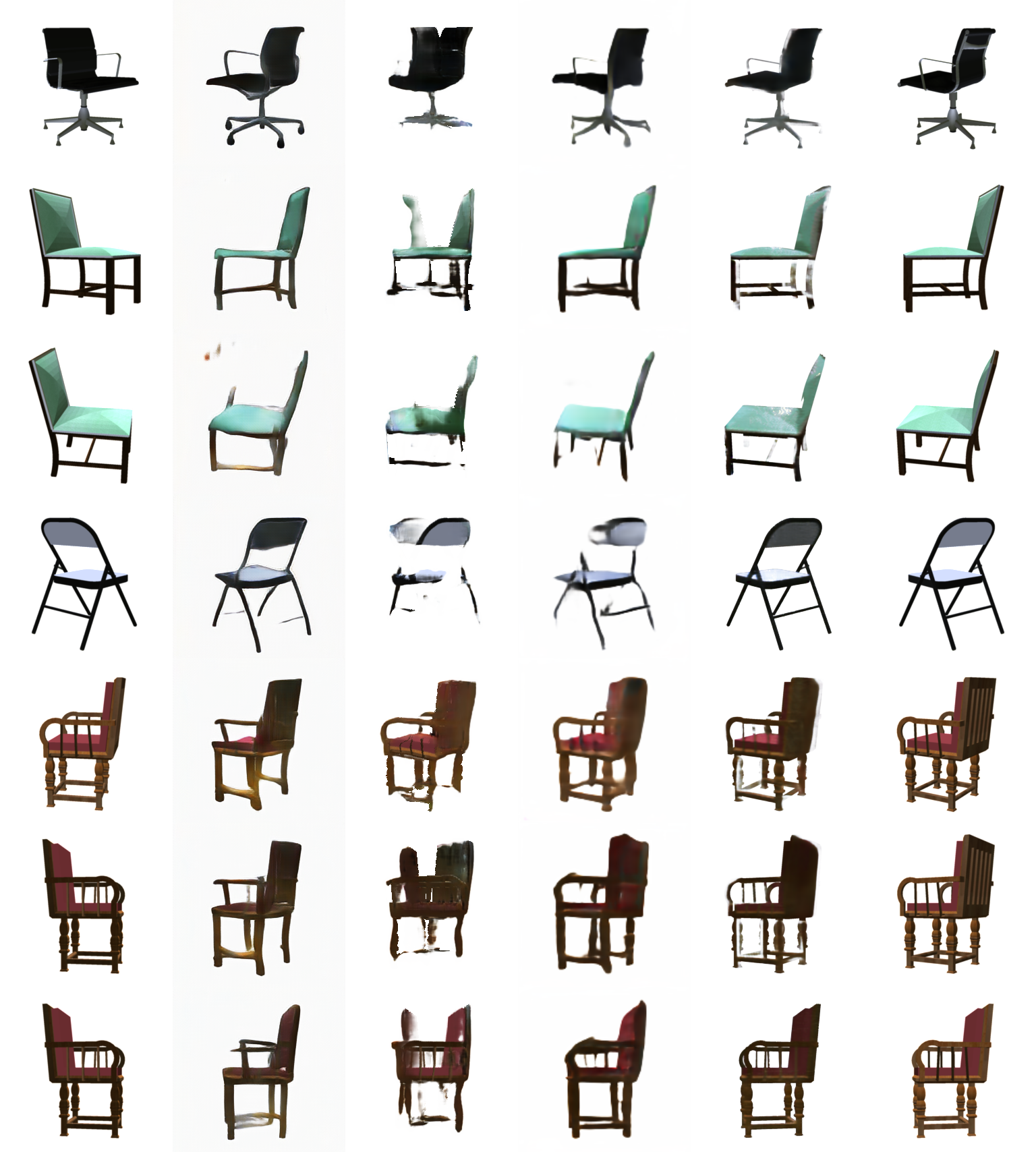}};
        \node at (-5.5,8.0) {Input};
        \node at (-3.3,8.0) {TVSN};
        \node at (-1.1,8.0) {M2NV};
        \node at ( 1.1,8.0) {TBN};
        \node at ( 3.3,8.0) {Ours};
        \node at ( 5.5,8.0) {Target};
    \end{tikzpicture}
    \caption{Qualitative examples of chairs generated by our point-cloud-based method and other related work. Our method retains better the objects' geometrical structure and detailed textures, \eg row 4 and 7. }
    \label{fig:sup_chair1}
\end{figure}

\begin{figure}
    \centering
        \begin{tikzpicture}
        \node{\includegraphics[width=\textwidth]{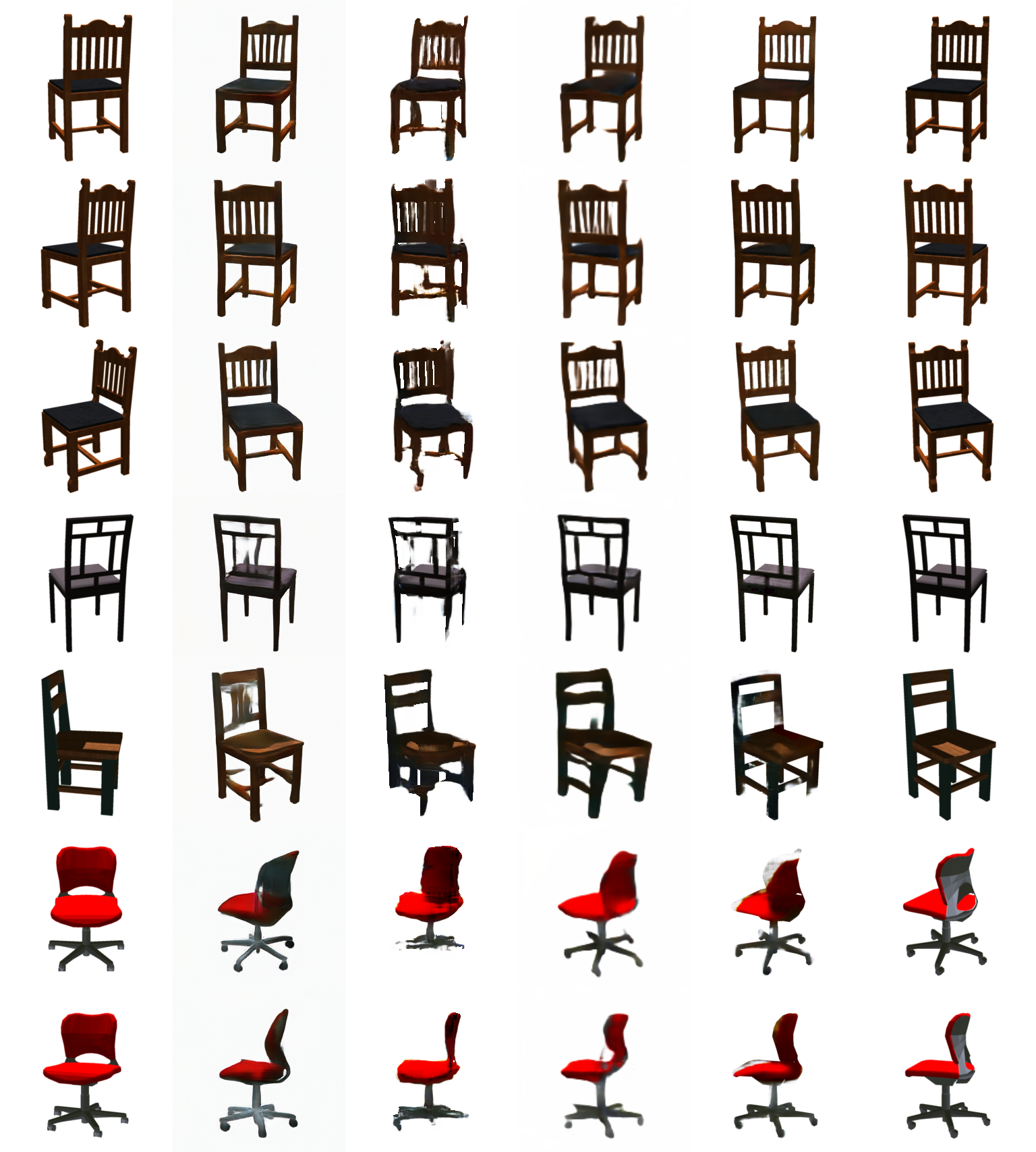}};
        \node at (-5.5,8.0) {Input};
        \node at (-3.3,8.0) {TVSN};
        \node at (-1.1,8.0) {M2NV};
        \node at ( 1.1,8.0) {TBN};
        \node at ( 3.3,8.0) {Ours};
        \node at ( 5.5,8.0) {Target};
    \end{tikzpicture}
    \caption{Qualitative examples of chairs generated by our point-cloud-based method and other related work. Our method retains better the objects' geometrical structure and detailed textures, \eg row 3 and 4. }
    \label{fig:sup_chair2}
\end{figure}

\begin{figure}
    \centering
        \begin{tikzpicture}
        \node{\includegraphics[width=\textwidth]{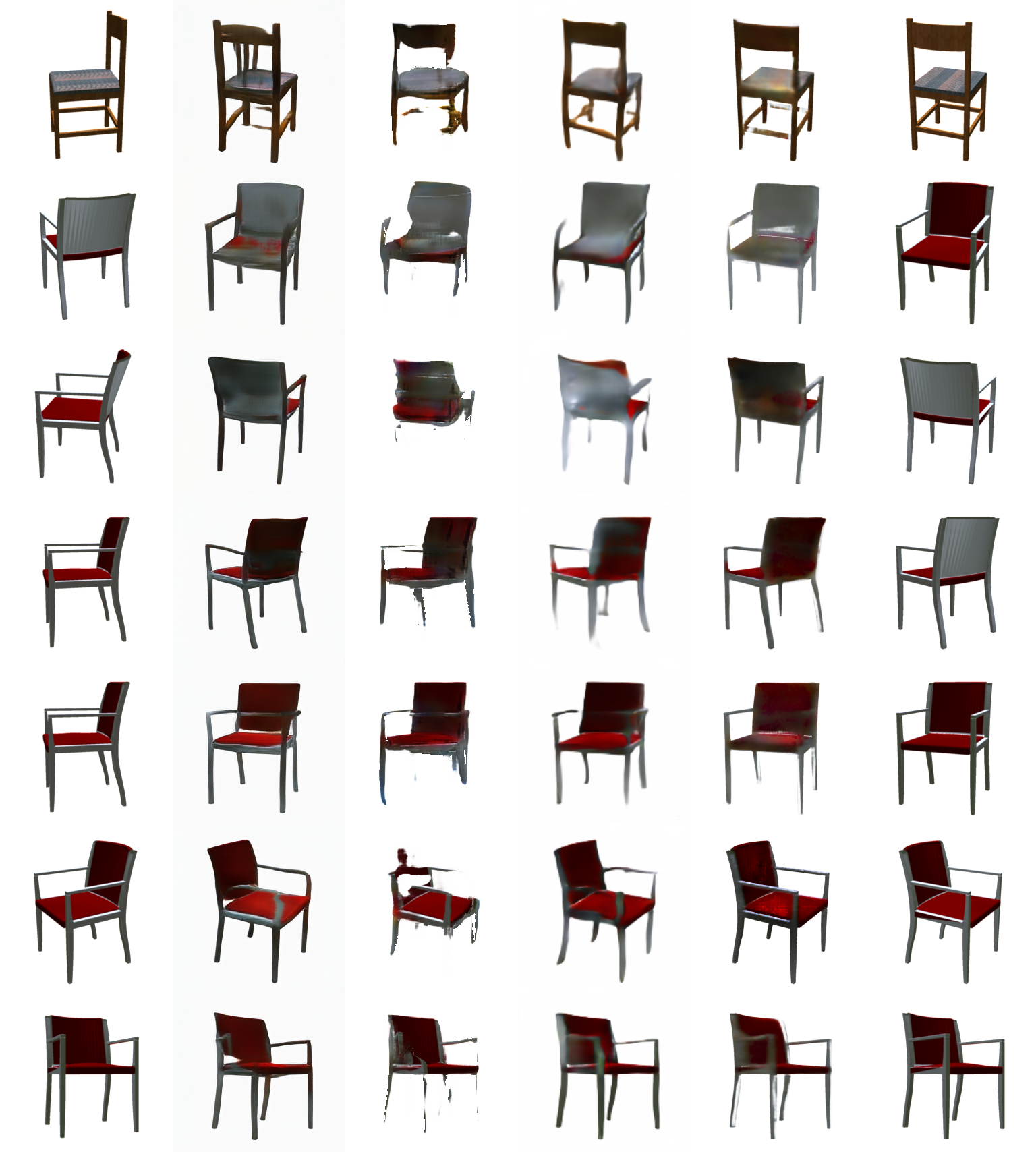}};
        \node at (-5.5,8.0) {Input};
        \node at (-3.3,8.0) {TVSN};
        \node at (-1.1,8.0) {M2NV};
        \node at ( 1.1,8.0) {TBN};
        \node at ( 3.3,8.0) {Ours};
        \node at ( 5.5,8.0) {Target};
    \end{tikzpicture}
    \caption{Qualitative examples of chairs generated by our point-cloud-based method and other related work. Our method retain better the objects' geometrical structure and detailed textures, \eg row 6 and 7. }
    \label{fig:sup_chair3}
\end{figure}

In the supplementary materials a more elaborate qualitative comparison is provided between the proposed method and other state-of-the-art methods.
For this the synthesized target views for 36 car images are shown in Figure~\ref{fig:sup_car1}-~\ref{fig:sup_car3} and the syntesized target views for 21 chair images are shown 
in Figure~\ref{fig:sup_chair1}-~\ref{fig:sup_chair3}.

From the results we observe that in line with the overall quality metrics (Table 2 of the main paper), our method synthesize in general higher quality (better geometrical shape and matching texture) compared to the other methods.

A few observations:
\begin{itemize}
    \item Observe that TVSN, M2NV and the proposed method all have a similar inter-connection network architecture, in contrast to TBN. The inter-connection allow for explicit use of the input image pixels in constructing the generated views, and thus these models can retain the object textures in the generated views. 
    Examples include row 4, 5 of Figure~\ref{fig:sup_car2} and row 2, 5, 10 of Figure~\ref{fig:sup_car3}, where the specific color patterns or texts on the input views are retained in the generated views.
    \item Note also row 4 of Figure~\ref{fig:sup_car2}, which is indeed a failure, yet legitimate, case. The target is posed at an extreme angle with the input, but the unseen back of the truck has a different texture/color. The methods based on the assumption of object symmetry to make an "educated" guess of the unseen view, thus fail when the object texture does not follow such assumption. 
    \item The main difference of our method to that of other state-of-the-arts is the explict use of object geometry in reasoning of occlusion, symmetry and in generating new views. TVSN and M2NV use occlusion and symmetry in creating annotated data and in training the network to predict the coarse view, while the proposed method impose such assumption directly on the coarse view. This has the benefit when the target pose is close to the symmetric pose of the input. Examples can be seen in the generated chair images, specifically row 2-7 of Figure~\ref{fig:sup_chair1} and row 2-4 of Figure~\ref{fig:sup_chair2}. Despite the intricate structure of chairs, these examples standout in quality compared to other methods.
\end{itemize}

\newpage

\newpage

\newpage

\newpage

\end{document}